\documentclass[11pt]{article}

\usepackage[preprint]{acl}

\usepackage{times}
\usepackage{latexsym}

\usepackage[T1]{fontenc}

\usepackage[utf8]{inputenc}

\usepackage{microtype}

\usepackage{inconsolata}

\usepackage{graphicx}

\newcommand{\model}{VideoLatent}

\usepackage{xspace}
\makeatletter
\DeclareRobustCommand\onedot{\futurelet\@let@token\@onedot}
\def\@onedot{\ifx\@let@token.\else.\null\fi\xspace}
\def\eg{\emph{e.g}\onedot} 
\def\ie{\emph{i.e}\onedot}

\makeatother
\usepackage{booktabs}
\usepackage{multirow}
\usepackage{array}
\usepackage[table]{xcolor}
\definecolor{LightGray}{gray}{0.97}
\definecolor{LightBlue}{rgb}{0.97,0.985,1.0}
\definecolor{LightGreen}{HTML}{F5FFFA}
\definecolor{StandardMLLMColor}{rgb}{0.97,0.985,1.0}
\definecolor{LatentMLLMColor}{HTML}{F5FFFA}
\definecolor{OurColor}{HTML}{F5FFFA}
\definecolor{grey}{gray}{0.95}
\definecolor{deepgrey}{gray}{0.5} 
\usepackage{amsmath}
\usepackage{amssymb}

\usepackage{pifont}
\usepackage{makecell}
 \usepackage{float}
 
\title{\model{}: Video-Language Learning via Latent Self-Forcing}

\author{
Zi-Yuan Hu$^{1,2}$\thanks{Work was done during an internship at Weitu AI.} \qquad
Zicong Tang$^{2*}$ \qquad
Shijia Huang$^{2}$ \qquad
Yanyang Li$^{2}$ \\
\textbf{Michael R. Lyu}$^1$ \qquad
\textbf{Liwei Wang}$^{1}$ \\
$^1$The Chinese University of Hong Kong \quad 
$^2$Weitu AI 
}

\begin{document}
\maketitle
\begin{abstract}
Recent advancements in chain-of-thought (CoT) reasoning have shown promise in enhancing video understanding and reasoning capabilities of multimodal large language models (MLLMs). 
However, existing CoT-based MLLMs require labor-intensive CoT annotations and incur substantial training and inference overhead.
While visual latent reasoning has emerged as a more efficient alternative, existing methods primarily focus on image tasks and heavily rely on additional supervision signals for visual latent generation (\eg, CoT traces, auxiliary images, or fine-grained annotations), limiting their scalability and transferability to video tasks.
To bridge this gap, we introduce \textbf{\model{}}, a novel MLLM equipped with a latent injection module tailored for video understanding and reasoning.
Specifically, \model{} learns to perform visual latent reasoning using a new latent self-forcing training paradigm, which comprises latent alignment and latent diversity objectives, and relies solely on standard video-question-answer triplets.
Extensive experiments across 14 benchmarks demonstrate that our model consistently outperforms existing standard and latent MLLMs on general video understanding and complex video reasoning.
Compared with Video-R1, our \model{} achieves superior computational efficiency, reducing training/inference overhead by $\sim$6$\times$/$\sim$68$\times$.
Moreover, experiments demonstrate that our method has strong generalizability to different MLLM backbones and different model scales.

\end{abstract}

\section{Introduction}
\label{sec:intro}
Recently, large language models (LLMs)~\cite{vaswani2017attention,llama_tech,DeepSeek-R1,gpt5,qwen3} have substantially enhanced complex reasoning capability through chain-of-thought (CoT) reasoning~\cite{cot}.
Building upon these advances, multimodal learning efforts have increasingly focused on equipping multimodal large language models (MLLMs) with CoT reasoning to enhance multimodal understanding and reasoning~\cite{gpto1,visionr1,mvot,VisualPlanning,Render-of-Thought,L2V-CoT,icot}.
Particularly in the video-language domain, numerous studies have been proposed to investigate the effectiveness of CoT reasoning in improving both general video understanding and complex video reasoning capabilities~\cite{videor1,videorft,open-o3-video,onethinker,Video-of-Thought,Chain-of-Frames,videomind,VideoEspresso}.

\begin{figure}[t]
    \centering
    \includegraphics[width=\linewidth]{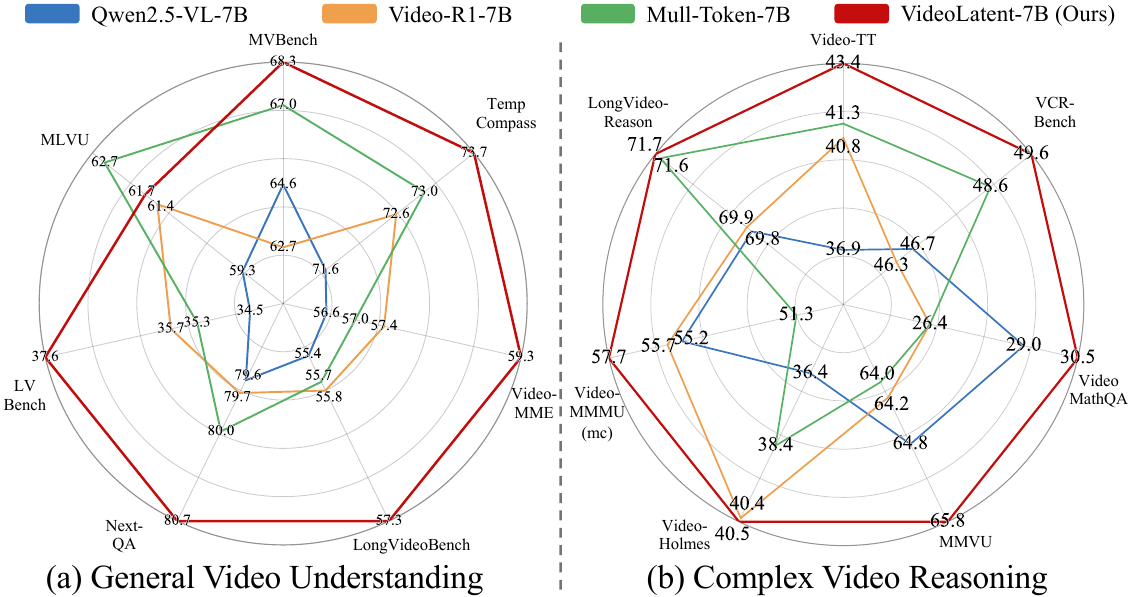}
    \caption{Our \model{}-7B consistently outperforms existing standard and latent MLLMs across fourteen benchmarks under the same experimental settings (see Tab.~\ref{tab:main_exp} and Tab.~\ref{tab:additional_exp} for more details), covering both general video understanding and complex video reasoning.}
    \label{fig:radar}
\end{figure}

Despite their improved performance and interpretability, existing CoT-based video reasoning approaches exhibit two critical limitations: 
(1) Their explicit reasoning traces may suffer from hallucination and overthinking~\cite{videoautor1,rethinking-cot}, and may not consistently improve performance on both general video understanding and complex video reasoning benchmarks (as demonstrated in Tab.~\ref{tab:main_exp}); 
and (2) Their training incurs substantial computational and memory overhead, and often requires labor-intensive CoT annotations~\cite{videor1,onethinker,open-o3-video}, limiting training scalability and inference efficiency.

To address these limitations, visual latent reasoning~\cite{mirage,lvr,Mull-Tokens,latent-survey} has emerged as a promising paradigm to perform implicit reasoning beyond language~\cite{coconut,softcot}, enabling more efficient and flexible reasoning. 
Specifically, it alternates between language-space and latent-space reasoning, with the latter generating continuous latent thoughts rather than discrete textual ones.
However, existing methods mainly focus on image understanding and reasoning~\cite{mirage,lvr,monet}, leaving visual latent reasoning for video tasks underexplored.
Moreover, they largely rely on additional, image-centric supervision signals for visual latent generation (\eg, CoT traces~\cite{heima}, helper images~\cite{mirage}, pretrained vision foundation models~\cite{covt}, bounding boxes~\cite{lvr}, and other fine-grained annotations~\cite{monet,laser}), limiting scalability and transferability to video reasoning.

\begin{figure*}[t]
    \centering
    \includegraphics[width=\linewidth]{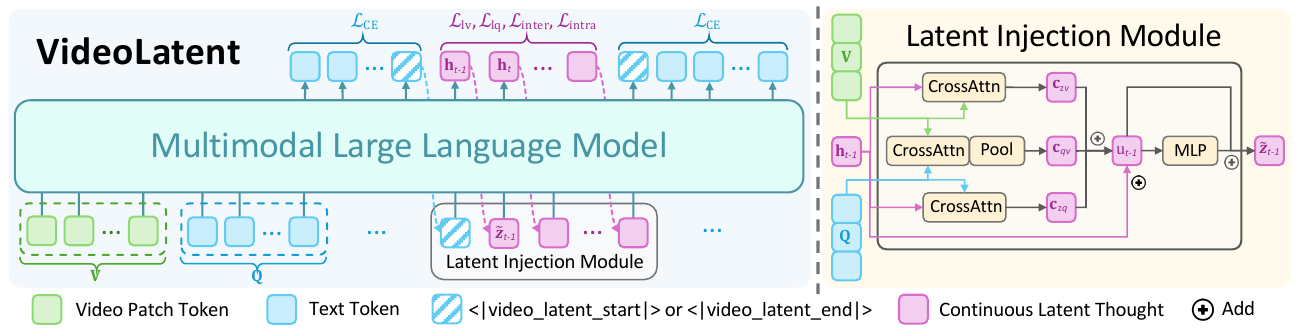}
    \caption{\textbf{Overview of \model{}}. Given an input video and a text question, our \model{} learns to perform visual latent reasoning (see Sec.~\ref{sec:our_model}) using our proposed latent self-forcing training paradigm (see Sec.~\ref{sec:our_training}). 
    Specifically, we introduce a latent injection module to prevent self-generated latent thoughts from drifting away from the video and question context.
    Furthermore, our latent self-forcing covers both latent alignment and latent diversity objectives to enhance video-language learning.
    Detailed symbol definitions are provided in Sec.~\ref{sec:method}.}
    \label{fig:overview}
\end{figure*}

In this paper, we propose \textbf{\model{}}, a novel MLLM framework focused on visual latent reasoning for video understanding and reasoning tasks, mitigating the limitations of both standard and latent MLLMs.
As in Fig.~\ref{fig:overview}, \model{} learns to perform visual latent reasoning through a new \textbf{latent injection module} and \textbf{latent self-forcing training paradigm}.
Specifically, the latent injection module is designed to prevent the self-generated latent thoughts from drifting away from the video and question context.
Latent self-forcing further enhances video-language learning via two complementary objectives (\ie, latent alignment and latent diversity), while relying solely on video-question-answer triplets without requiring any additional supervision for visual latent generation.

\begin{figure}[t]
    \centering
    \includegraphics[width=\linewidth]{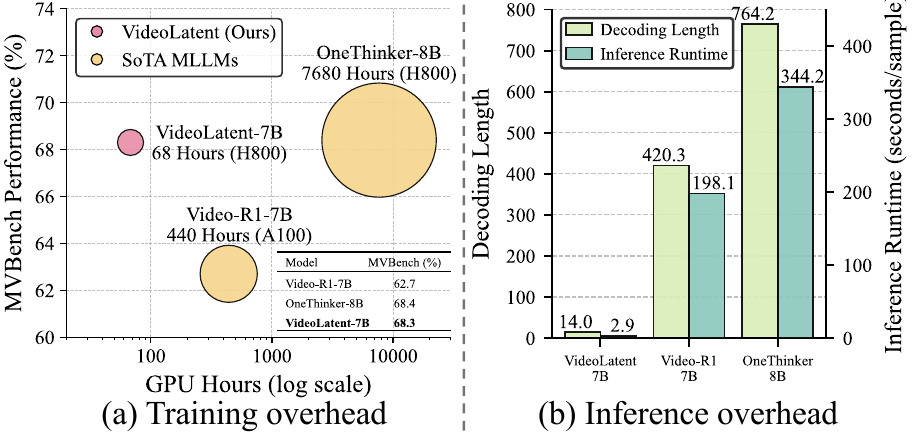}
    \caption{Our \model{} achieves stronger or comparable performance with existing CoT-based MLLMs, while achieving superior computational efficiency.}
    \label{fig:overhead}
\end{figure}

To ensure a comprehensive evaluation, we conduct experiments on fourteen video-language benchmarks, covering both general video understanding and complex video reasoning.
Our experiments yield several \textbf{key findings}: 
(1) \model{} consistently outperforms both standard and latent MLLMs across diverse benchmarks and experimental settings (\eg, Tab.~\ref{tab:main_exp} and Tab.~\ref{tab:additional_exp}), demonstrating the \textbf{effectiveness} of our proposed method;
(2) Our method exhibits strong \textbf{generalizability} to different MLLM backbones and different model sizes (\eg, Tab.~\ref{tab:main_exp2} and Tab.~\ref{tab:exp_3b});
(3) Our \model{} achieves superior computational \textbf{efficiency} compared with CoT-based MLLMs (\eg, Fig.~\ref{fig:overhead});
and (4) Visualization and ablation studies further demonstrate the effectiveness of our design. 

Our contributions are summarized as follows: 
\begin{enumerate}
    \item We introduce \model{}, a novel MLLM equipped with a latent injection module to enhance video-language learning.
    \item We propose a new latent self-forcing training paradigm that enables \model{} to perform visual latent reasoning, relying solely on standard video-question-answer triplets.
    \item Extensive experiments conducted on fourteen benchmarks, covering both general video understanding and complex video reasoning, demonstrate the effectiveness, efficiency, and generalizability of our proposed method.
\end{enumerate}

\section{Related Work}
\noindent \textbf{MLLLMs for Video Understanding and Reasoning.}
With the rapid evolution of large language models (LLMs)~\cite{vaswani2017attention,llama_tech,DeepSeek-R1,gpt,gpt4_tech,gpt5,gemini,gemini1.5,qwen2.5,qwen3}, multimodal large language models (MLLMs) have witnessed remarkable progress in the vision-language domain~\cite{llava,llava1_5, Llava_next,LLaVA-OneVision,qwen_vl, Mini_Gemini24, blip_23,vila,LLaVA-Video,mPLUG-Owl3,vlpet,videochatgpt,Molmo,InternVL3.5,NVILA,InstructBLIP,videossr,v-jepa2,vl-jepa}. 
For example, existing MLLMs have demonstrated strong performance on general video understanding benchmarks~\cite{mmvu,mvp,aotbench,Vinoground,PerceptionTest,TemporalBench,TVBench,TOMATO,MotionBench,lvbench}.
To further enhance complex video reasoning ability~\cite{videommmu,vsibench,vsi-super,longvideoreason,CG-Bench,Video-Holmes,mmrvbench,VRBench}, recent efforts have explored incorporating explicit chain-of-thought (CoT) reasoning into MLLMs~\cite{cot,autocot,mmcot,visualtable,ccot,ddcot,longvila-r1,ego-r1,Video-RTS}.

Specifically, Video-R1~\cite{videor1} proposes T-GRPO (\ie, a variant of GRPO~\cite{DeepSeekMath}) to efficiently enable R1-style~\cite{DeepSeek-R1} reasoning over videos, while VideoRFT~\cite{videorft} strengthens the alignment between textual reasoning and visual evidence via a semantic-consistency reward.
VideoChat-R1.5~\cite{videochatr1.5} further improves multimodal reasoning by iteratively refining its focus on fine-grained regions.
OneThinker~\cite{onethinker} optimizes multimodal reasoning by employing a new EMA-GRPO algorithm, whereas Open-o3-Video~\cite{open-o3-video} learns to integrate explicit spatio-temporal evidence into the video reasoning process.
Concurrently, VideoAuto-R1~\cite{videoautor1} introduces a reason-when-necessary strategy to facilitate dynamic CoT reasoning, while still incurring significant training overhead (\ie, 1120 H100 GPU hours).

Despite their promising performance, existing CoT-based video reasoning approaches primarily rely on explicit CoT reasoning and additional labor-intensive CoT supervision~\cite{videor1}, resulting in substantial computational overhead and limited training scalability.
In contrast, we propose \model{}, which introduces a new latent injection module and latent self-forcing to enhance video-language learning without explicit CoT annotations, while maintaining strong performance and high computational efficiency.

\smallskip
\noindent \textbf{Visual Latent Reasoning.}
Latent reasoning has demonstrated strong effectiveness in NLP, enabling LLMs to perform reasoning beyond the constraints of language~\cite{Pause-pretraining,implicit-cot,codi,compressed-cot,softcot-plusplus,latent-agent,latent-survey}.
For example, \cite{coconut,softcot} learn to switch between language and latent reasoning modes.
Instead of generating explicit textual thoughts, latent reasoning performs implicit reasoning in continuous latent spaces, enabling more efficient and flexible reasoning.
Recently, latent reasoning has been extended to a wide range of domains, including image reasoning~\cite{lavit,vedas,dmlr,heima,Sketch-in-Latents}, 3D reasoning~\cite{3dthinker}, image generation~\cite{DBLP:journals/corr/abs-2602-02227}, robotics~\cite{DBLP:journals/corr/abs-2511-19859,last0}, and autonomous driving~\cite{last-vla}, with some works using foundation models (\eg, Diffusion~\cite{cocova,DBLP:journals/corr/abs-2602-00574,LatentSketchpad,DBLP:conf/aaai/HeLCW24} and DINO~\cite{valr}) to supervise visual latent generation.

In particular, visual latent reasoning has emerged as a promising direction for enhancing the multimodal understanding and reasoning capabilities of MLLMs~\cite{Aurora-perception,ilvr}. 
For instance, 
Mirage~\cite{mirage} learns to reconstruct the image embeddings of helper images as visual latent representations, while ILVR~\cite{ilvr} instead learns to selectively extract context-relevant visual signals from helper images.
Furthermore, CoVT~\cite{covt} leverages multiple pretrained vision foundation models~\cite{sam,depth,PIDINet,dinov2} to supervise the visual latent generation.
Concurrently, LVR~\cite{lvr} learns to generate fine-grained visual latent representations by reconstructing the embeddings of cropped image regions based on bounding box annotations, whereas Monet~\cite{monet} further incorporates diverse fine-grained supervision signals (\eg, cropping, grounding, highlighting) into visual latent reasoning.
Laser~\cite{laser} learns to align visual latent representations with cognitive scanpaths annotated by GPT-4o~\cite{gpt4o}, while Mull-Tokens~\cite{Mull-Tokens} enables modality-agnostic latent reasoning using additional annotations, including both textual CoT traces and subgoal images. 

Despite recent progress, visual latent reasoning for video tasks remains underexplored and faces significant scalability challenges. 
Specifically, existing methods primarily rely on additional, image-centric supervision signals for visual latent generation (\eg, CoT traces~\cite{heima}, helper images~\cite{mirage}, pretrained vision foundation models~\cite{covt}, fine-grained annotations~\cite{lvr,monet,laser}), limiting both training scalability and transferability to video tasks.
In contrast, we propose a latent injection module and a latent self-forcing training paradigm tailored for visual latent reasoning, without requiring additional latent supervision.

\section{Methodology}
\label{sec:method}
In this section, we first describe the formulation of standard MLLMs in Sec.~\ref{sec:preliminary}. 
To enhance video-language learning, we introduce \model{}, a new latent MLLM equipped with a latent injection module in Sec.~\ref{sec:our_model}. 
We further propose a new latent self-forcing training paradigm (Sec.~\ref{sec:our_training}) to enable effective video latent reasoning.

\subsection{Preliminary: Standard MLLM}
\label{sec:preliminary}
A standard MLLM typically consists of a vision encoder, an LLM $\mathcal{M}_\theta$, and a modality projector.
Given an input video $\mathcal{V} = \{v_f\}_{f=1}^{F}$ with $F$ frames and a text question $\mathcal{T}$ 
(system and instruction prompts are omitted for simplicity), 
the model first extracts a sequence of visual patch tokens
$\mathbf{V} = \{\mathbf{v}_{f,k}\}_{f=1,k=1}^{F,N_p}$,
where $\mathbf{v}_{f,k} \in \mathbb{R}^{d}$ denotes the $k$-th projected patch token of the $f$-th video frame, $d$ is the hidden dimension, and $N_p$ is the number of patch tokens per frame.
The text input is encoded as a sequence of text embeddings
$\mathbf{Q} = \{\mathbf{q}_i\}_{i=1}^{N_q}$,
where $\mathbf{q}_i \in \mathbb{R}^{d}$ denotes the $i$-th text token embedding. 
Subsequently, the model generates an output token sequence
$\mathbf{Y} = \{y_t\}_{t=1}^{N_y}$ in an autoregressive manner:
\begin{equation}
\resizebox{0.85\hsize}{!}{
$
P(\mathbf{Y} \mid \mathbf{V}, \mathbf{Q}; \theta) = \prod_{t=1}^{N_y}
P(y_t \mid y_{<t}, \mathbf{V}, \mathbf{Q}; \theta),
$
}
\end{equation}
where $y_{<t} = \{y_1, \dots, y_{t-1}\}$ denotes previously generated tokens. 
For brevity, we denote the next-token distribution $P(y_t \mid y_{<t}, \mathbf{V}, \mathbf{Q}; \theta)$ as $\mathbf{p}_t$.
At each decoding step $t$, the LLM backbone computes the last-layer hidden state $\mathbf{h}_t \in \mathbb{R}^{d}$, next-token distribution $\mathbf{p}_t$, and next predicted token $y_t$ as follows (assuming greedy decoding for simplicity):
\begin{equation}
\label{eq:h_t}
\resizebox{0.55\hsize}{!}{
$
\begin{aligned}
\mathbf{h}_t &= \mathcal{M}_\theta([\mathbf{V}, \mathbf{Q}, \mathbf{E}(y_{<t})]), \\
\mathbf{p}_t &= \mathrm{Softmax}(\mathbf{W}_o \mathbf{h}_t), \\
y_t &= \arg\max_{y \in \mathcal{Y}} \mathbf{p}_t(y),
\end{aligned}
$
}
\end{equation}
where $[\cdot]$ denotes sequence concatenation, $\mathbf{E}(\cdot)$ is the text embedding lookup function, $\mathbf{W}_o \in \mathbb{R}^{|\mathcal{Y}| \times d}$ denotes the MLLM's language head, and $|\mathcal{Y}|$ denotes the vocabulary size.

\subsection{Our Model: \model{} }
\label{sec:our_model}
In this section, we detail the design of our \model{}, which comprises a latent injection module and learns to perform visual latent reasoning via our proposed latent self-forcing, as shown in Fig.~\ref{fig:overview}.

\smallskip
\noindent \textbf{\model{} Overview.}
Our \model{} extends standard MLLMs by augmenting the output space to include both discrete text tokens and continuous latent thoughts (\ie, the last-layer hidden state $\mathbf{h}_t$).
Specifically, \model{} learns to switch between a language mode and a latent mode, with the former serving as the default, as in standard MLLMs.

At a given decoding step, \model{} enters the latent mode upon predicting a special token \texttt{<|video\_latent\_start|>}, and performs visual latent reasoning in the subsequent decoding steps.
During the latent mode, our \model{} directly generates the last-layer hidden state $\mathbf{h}_{t}$ as the predicted continuous latent thought, instead of generating a discrete text token $y_{t}$. 
\model{} reverts to the language mode either when the generated $\mathbf{h}_{t}$ is mapped to another special token \texttt{<|video\_latent\_end|>} via the language head (Eq.~\ref{eq:h_t}), or when a predefined number $N_{\text{max}}$ of latent reasoning steps is reached.
Assuming that the latent mode is triggered at step $t_s$ and lasts for $N_h$ steps, we denote the sequence of generated continuous latent thoughts as $\mathbf{H} = \{\mathbf{h}_t\}_{t=t_s+1}^{t_s+N_h}$. 
Consequently, the hybrid text-latent output sequence $\mathbf{Y}_h$ can be formulated as:
\begin{equation}
\resizebox{0.55\hsize}{!}{
$
\mathbf{Y}_h = [ y_{\le t_s}, \mathbf{H}, y_{> t_s+N_h} ],$
}
\end{equation}
where $y_{\le t_s}$ and $y_{> t_s+N_h}$ denote the discrete text token sequences before and after the latent mode. 
In particular, $y_{\le t_s}$ ends with the \texttt{<|video\_latent\_start|>} token, and $y_{> t_s+N_h}$ begins with the \texttt{<|video\_latent\_end|>} token.

\smallskip
\noindent \textbf{Latent Injection Module.}
During the latent mode, to generate the continuous latent thought $\mathbf{h}_t$ at step $t \in [t_s+1, t_s+N_h]$, we define the raw input representation $\mathbf{z}_{t-1} \in \mathbb{R}^{d}$ fed to our model as:
\begin{equation}
\resizebox{0.85\hsize}{!}{
$
\mathbf{z}_{t-1} = 
\begin{cases} 
\mathbf{E}(y_{t_s}), & \text{if } t = t_s + 1, \\
\mathbf{h}_{t-1}, & \text{if } t_s + 1 < t \le t_s + N_h.
\end{cases}
$
}
\end{equation}

To prevent the self-generated latent thoughts from drifting away from the input video and question context, we introduce a latent injection module $\mathcal{F}_\phi$ tailored for enhancing video-language learning
As shown in Fig.~\ref{fig:overview}, the latent injection module dynamically injects the video features $\mathbf{V}$ and text question features $\mathbf{Q}$ into $\mathbf{z}_{t-1}$:
\begin{equation}
\resizebox{0.5\hsize}{!}{
$
\tilde{\mathbf{z}}_{t-1} = \mathcal{F}_\phi(\mathbf{z}_{t-1}, \mathbf{V}, \mathbf{Q}),
$
}
\end{equation}
where $\tilde{\mathbf{z}}_{t-1}  \in \mathbb{R}^{d}$ denotes the ultimate context-aware representation feed into our \model{}.

Specifically, $\mathcal{F}_\phi$ first obtains the aggregated semantic information $\mathbf{u}_{t-1} \in \mathbb{R}^{d}$ from three complementary perspectives (\ie, latent-to-video, latent-to-question, question-to-video) via cross attention:
\begin{equation}
\resizebox{0.7\hsize}{!}{
$
\begin{aligned}
\mathbf{c}_{\text{z}v} &= \mathrm{CrossAttn}(\mathbf{z}_{t-1}, \mathbf{V}) \in \mathbb{R}^{d}, \\
\mathbf{c}_{\text{z}q} &= \mathrm{CrossAttn}(\mathbf{z}_{t-1}, \mathbf{Q}) \in \mathbb{R}^{d}, \\
\mathbf{c}_{qv} &= \mathrm{Pool}(\mathrm{CrossAttn}(\mathbf{Q}, \mathbf{V})) \in \mathbb{R}^{d}, \\
\mathbf{u}_{t-1} & = \mathbf{z}_{t-1} + \mathbf{c}_{\text{z}v} + \mathbf{c}_{\text{z}q} + \mathbf{c}_{qv}  \in \mathbb{R}^{d},
\end{aligned}
$
}
\end{equation}
where $\mathrm{Pool(\cdot)}$ denotes mean pooling along the sequence dimension. $\mathcal{F}_\phi$ then produces $\tilde{\mathbf{z}}_{t-1}$ via an MLP with a residual connection:
\begin{equation}
\resizebox{0.55\hsize}{!}{
$
\tilde{\mathbf{z}}_{t-1} = \mathbf{u}_{t-1} + \mathrm{MLP}(\mathbf{u}_{t-1}).
$
}
\end{equation}

\subsection{Our Training: Latent Self-Forcing}
\label{sec:our_training}
Prior works on visual latent reasoning mainly focus on optimizing the continuous latent thoughts $\mathbf{H} = \{\mathbf{h}_t\}_{t=t_s+1}^{t_s+N_h}$ using additional, image-centric supervision signals for visual latent generation (\eg, CoT traces~\cite{heima}, helper images~\cite{mirage}, pretrained vision foundation models~\cite{covt}, bounding boxes~\cite{lvr}, and other fine-grained annotations~\cite{monet,laser}), which limits their scalability and transferability to video reasoning.

In contrast, we introduce a latent self-forcing mechanism specifically designed to enhance video-language learning, without requiring additional annotations (\ie, training relies solely on video–question–answer triplets). 
Specifically, latent self-forcing comprises latent alignment and latent diversity objectives to supervise the generation of continuous latent thoughts. 

\smallskip
\noindent \textbf{Latent Alignment.} 
To prevent the self-generated continuous latent thoughts from drifting away from the visual and linguistic context, we explicitly align $\mathbf{H}$ with $\mathbf{V}$ and $\mathbf{Q}$ via contrastive learning, where we follow~\cite {moco} to maintain dynamic memory banks with a queue size of $K$ to store negative samples (\eg, videos, questions, generated latent thoughts) from preceding batches. 

For \textit{Latent-Video Alignment}, we first apply adaptive pooling to $\mathbf{V}$ to obtain $\bar{\mathbf{V}} \in \mathbb{R}^{N_h \times d}$. We then flatten and $L_2$-normalize both $\mathbf{H}$ and $\bar{\mathbf{V}}$ into $\tilde{\mathbf{h}} \in \mathbb{R}^{N_h d}$ and $\tilde{\mathbf{v}} \in \mathbb{R}^{N_h d}$. The latent-video alignment loss $\mathcal{L}_{\text{lv}}$ is formulated as:
\begin{equation}
\label{eq:loss_lv}
\resizebox{0.7\hsize}{!}{
$
\mathcal{L}_{\text{lv}} = -\log \frac{e^{\mathrm{sim}(\tilde{\mathbf{h}}, \tilde{\mathbf{v}}) / \tau}}{e^{\mathrm{sim}(\tilde{\mathbf{h}}, \tilde{\mathbf{v}}) / \tau} + \sum_{k=1}^{K} e^{\mathrm{sim}(\tilde{\mathbf{h}}, \tilde{\mathbf{v}}^-_k) / \tau}},$
}
\end{equation}
where $\mathrm{sim}(\cdot)$ denotes the cosine similarity, $\tau$ is the temperature hyperparameter, and $\tilde{\mathbf{v}}^-_k$ represents the negative video samples from the memory bank.

For \textit{Latent-Question Alignment}, we first employ mean pooling to $\mathbf{Q}$ to extract the global question representation $\bar{\mathbf{q}} \in \mathbb{R}^d$. To ensure the question-awareness of each latent thought, the latent-question alignment loss $\mathcal{L}_{\text{lq}}$ is defined as:
\begin{equation}
\label{eq:loss_lq}
\resizebox{0.87\hsize}{!}{
$
\mathcal{L}_{\text{lq}} = -\frac{1}{N_h} \sum_{\mathbf{h}_t \in \mathbf{H}} \log \frac{e^{\mathrm{sim}(\mathbf{h}_t, \bar{\mathbf{q}}) / \tau}}{e^{\mathrm{sim}(\mathbf{h}_t, \bar{\mathbf{q}}) / \tau} + \sum_{k=1}^{K} e^{\mathrm{sim}(\mathbf{h}_t, \mathbf{q}^-_k) / \tau}},
$
}
\end{equation}
where $\mathbf{q}^-_k$ represents the negative question samples.

\smallskip
\noindent \textbf{Latent Diversity.} 
To avoid latent representation collapse, we adopt contrastive learning for \textit{Inter-Latent Diversity}.
Similar to Eq.~\ref{eq:loss_lv}, the inter-latent diversity loss is formulated as:
\begin{equation}
\label{eq:loss_inter}
\resizebox{0.7\hsize}{!}{
$
\mathcal{L}_{\text{inter}} = -\log \frac{e^{\mathrm{sim}(\tilde{\mathbf{h}}, \tilde{\mathbf{h}}) / \tau}}{e^{\mathrm{sim}(\tilde{\mathbf{h}}, \tilde{\mathbf{h}}) / \tau} + \sum_{k=1}^{K} e^{\mathrm{sim}(\tilde{\mathbf{h}}, \tilde{\mathbf{h}}^-_k) / \tau}},
$
}
\end{equation}
where $\tilde{\mathbf{h}}^-_k$ denotes the negative samples.

\begin{table*}[!t]
\centering
\resizebox{\linewidth}{!}{
\setlength{\tabcolsep}{1.4mm}
\renewcommand\arraystretch{1.3}
\begin{tabular}{lcccccccc}
\toprule
\multirow{2}{*}{\textbf{Model}} & \multirow{2}{*}{\textbf{Frames}} & \multicolumn{4}{c}{\textbf{General Video Understanding}} & \multicolumn{3}{c}{\textbf{Complex Video Reasoning}} \\ 
\cmidrule(lr){3-6} \cmidrule(lr){7-9}
      & & {MVBench} & {TempCompass} & {Video-MME} & {LongVideoBench} & {Video-TT} & {VCR-Bench} & {VideoMathQA} \\

\midrule
\rowcolor{StandardMLLMColor}
\multicolumn{9}{c}{\textit{Standard MLLMs}} \\
\rowcolor{StandardMLLMColor}
Qwen2.5-VL-7B  & 16 & 64.6 & 71.6 & 56.6 & 55.4 & 36.9 & 46.7 & \underline{29.0} \\
\rowcolor{StandardMLLMColor}
Qwen2.5-VL-7B (CoT) & 16 & 57.4$^\spadesuit$ & 72.2$^\spadesuit$ & 53.1$^\spadesuit$ & - & - & - & - \\
\rowcolor{StandardMLLMColor}
Video-R1-7B & 16 & 62.7$^\spadesuit$ & 72.6$^\spadesuit$ & 57.4$^\spadesuit$ & 55.8 & 40.8 & 46.3 & 26.4 \\
\rowcolor{StandardMLLMColor}
\rowcolor{LatentMLLMColor}
\multicolumn{9}{c}{\textit{Latent MLLMs}} \\
\rowcolor{LatentMLLMColor}
Monet-7B & 16 & 37.1 & 31.2 & 28.9 & 19.7 & 20.6 & 24.7 & 11.7 \\
\rowcolor{LatentMLLMColor}
LVR-7B & 16 & 64.6 & 72.0 & \underline{59.1} & \underline{56.4} & 37.0 & 45.3 & 26.9 \\
\rowcolor{LatentMLLMColor}
Mull-Token-7B & 16 & \underline{67.0} & \underline{73.0} & 57.0 & 55.7 & \underline{41.3} & \underline{48.6} & 26.4 \\
\rowcolor{OurColor}
\textbf{VideoLatent-7B (Ours)} & 16 & \textbf{68.3} & \textbf{73.7} & \textbf{59.3} & \textbf{57.3} & \textbf{43.4} & \textbf{49.6} & \textbf{30.5}  \\
\rowcolor{OurColor}
\multicolumn{2}{c}{\textit{$\Delta$ (vs. Qwen2.5-VL-7B)}} & \textit{+3.7} & \textit{+2.1} & \textit{+2.7} & \textit{+1.9} & \textit{+6.5} & \textit{+2.9} & \textit{+1.5} \\

\midrule
\rowcolor{StandardMLLMColor}
\multicolumn{9}{c}{\textit{Standard MLLMs}} \\
\rowcolor{StandardMLLMColor}
Qwen2.5-VL-7B  & 32 & 67.4 & 72.3 & 59.4 & \underline{58.8} & 37.9 & 50.4 & \underline{27.9} \\
\rowcolor{StandardMLLMColor}
Qwen2.5-VL-7B (CoT) & 32 & 59.0$^\spadesuit$ & 72.6$^\spadesuit$ & 56.6$^\spadesuit$ & - & - & - & - \\
\rowcolor{StandardMLLMColor}
Video-R1-7B & 32 & 63.9$^\spadesuit$ & 73.2$^\spadesuit$ & 59.3$^\spadesuit$ & 56.8 & \underline{41.8} & 46.9 & 24.3 \\
\rowcolor{StandardMLLMColor}
VideoRFT-7B & 32 & 62.1$^\spadesuit$ & \underline{73.7$^\spadesuit$} & 59.8$^\spadesuit$ & - & - & - & - \\
\rowcolor{StandardMLLMColor}
Open-o3-Video-7B & 32 & 66.0 & - & \underline{61.3} & \underline{58.8} & 39.3 & \underline{52.5} & 25.7 \\
\rowcolor{LatentMLLMColor}
\multicolumn{9}{c}{\textit{Latent MLLMs}} \\
\rowcolor{LatentMLLMColor}
Monet-7B & 32 &  38.9 & 34.4 & 33.2 & 18.3 & 20.5 & 25.5 & 14.0 \\
\rowcolor{LatentMLLMColor}
LVR-7B & 32 & 65.1 & 72.0 & 60.5 & \textbf{60.5} & 38.3 & 49.0 & 25.0 \\
\rowcolor{LatentMLLMColor}
Mull-Token-7B & 32 & \underline{68.5} & 73.6 & 59.5 & 57.2 & 40.8 & 50.6 & 24.8 \\
\rowcolor{OurColor}
\textbf{VideoLatent-7B (Ours)} & 32 & \textbf{69.4} & \textbf{74.1} & \textbf{61.4} & 58.3 & \textbf{44.4} & \textbf{53.1} & \textbf{30.0} \\
\rowcolor{OurColor}
\multicolumn{2}{c}{\textit{$\Delta$ (vs. Qwen2.5-VL-7B)}} & \textit{+2.0} & \textit{+1.8} & \textit{+2.0} & \textit{-0.5} & \textit{+6.6} & \textit{+2.7} & \textit{+2.1} \\

\midrule
\rowcolor{StandardMLLMColor}
\multicolumn{9}{c}{\textit{Standard MLLMs}} \\
\rowcolor{StandardMLLMColor}
Qwen2.5-VL-7B  & 64 &  67.1 & 72.3 & \underline{62.7} & \underline{59.8} & 38.7 & \underline{51.4} & \textbf{29.8} \\
\rowcolor{StandardMLLMColor}
Qwen2.5-VL-7B (CoT) & 64 &  59.2$^\spadesuit$ & 72.9$^\spadesuit$ & 59.6$^\spadesuit$ & - & - & - & - \\
\rowcolor{StandardMLLMColor}
Video-R1-7B & 64 & 64.8$^\spadesuit$ & 73.2$^\spadesuit$ & 61.4$^\spadesuit$ & 57.6 & \underline{41.8} & 50.8 & 25.2 \\
\rowcolor{StandardMLLMColor}
Open-o3-Video-7B & 64 & 65.8 & - & 61.4 & 59.7 & 39.7 & 49.2  & 25.0  \\
\rowcolor{LatentMLLMColor}
\multicolumn{9}{c}{\textit{Latent MLLMs}} \\
\rowcolor{LatentMLLMColor}
Monet-7B & 64 & 39.3 & 34.5 & 34.8 & 19.8 & 21.5 & 25.1 & 12.9 \\
\rowcolor{LatentMLLMColor}
LVR-7B & 64 &  65.5 & 72.0 & 62.0 & \textbf{60.1} & 39.3 & 48.0 & 28.8 \\
\rowcolor{LatentMLLMColor}
Mull-Token-7B & 64 & \underline{68.3} & \underline{73.6} & 62.1 & 58.3 & 41.1 & 50.6 & 25.0 \\
\rowcolor{OurColor}
\textbf{VideoLatent-7B (Ours)} & 64 & \textbf{69.2} & \textbf{74.2} & \textbf{63.8} & \textbf{60.1} & \textbf{43.9} & \textbf{52.5} & \underline{29.5} \\
\rowcolor{OurColor}
\multicolumn{2}{c}{\textit{$\Delta$ (vs. Qwen2.5-VL-7B)}} & \textit{+2.1} & \textit{+1.9} & \textit{+1.1} & \textit{+0.3} & \textit{+5.2} & \textit{+1.1} & \textit{-0.3} \\
\bottomrule
\end{tabular}
}
\caption{\textbf{Main experiments: comparisons with standard and latent MLLMs on general video understanding and complex video reasoning} under varying numbers of input frames. Our model consistently outperforms existing standard MLLMs and latent MLLMs and achieves the best or second-best performance on most benchmarks. 
$\spadesuit$: Results are borrowed from~\cite{videor1,videorft}, while others are produced under the same experimental setting (\eg, using a frame resolution of 256$\times$32$\times$32 pixels; see Sec.~\ref{sec:exp_setup} for more details).
\textbf{Bold} / \underline{Underlined}: Best / Second-best result.}
\label{tab:main_exp}
\end{table*}

For \textit{Intra-Latent Diversity}, we penalize the similarity between different latent thoughts within $\mathbf{H}$, which is defined as: 
\begin{equation}
\resizebox{0.8\hsize}{!}{
$
\mathcal{L}_{\text{intra}} = \frac{1}{N_h (N_h - 1)} \sum_{\mathbf{h}_i, \mathbf{h}_j \in \mathbf{H}, i \neq j} \left( \mathrm{sim}(\mathbf{h}_i, \mathbf{h}_j) \right)^2.
$
}
\end{equation}

\smallskip
\noindent \textbf{Overall Training Objectives.}
The overall training objectives combine the standard next token prediction loss $\mathcal{L}_{\text{CE}}$ for discrete text token prediction, along with our proposed latent-centric losses:
\begin{equation}
\label{eq:loss_total}
\resizebox{0.87\hsize}{!}{
$
\mathcal{L} = \mathcal{L}_{\text{CE}} + \lambda_{\text{cl}} \big( \mathcal{L}_{\text{lv}} + \mathcal{L}_{\text{lq}} + \mathcal{L}_{\text{inter}} \big) + \lambda_{\text{intra}} \mathcal{L}_{\text{intra}},
$
}
\end{equation}
where $\lambda_{\text{cl}}$ and $\lambda_{\text{intra}}$ are balancing coefficients.

\begin{table*}[t]
\centering
\resizebox{\linewidth}{!}{
\setlength{\tabcolsep}{1.4mm}
\renewcommand\arraystretch{1.3}
\begin{tabular}{lcccccccc}
\toprule
\multirow{2}{*}{\textbf{Model}} & \multirow{2}{*}{\textbf{Frames}} & \multicolumn{3}{c}{\textbf{General Video Understanding}} & \multicolumn{4}{c}{\textbf{Complex Video Reasoning}} \\ 
\cmidrule(lr){3-5} \cmidrule(lr){6-9}
      & & {Next-QA} & {LVBench} & {MLVU} & {MMVU} & {Video-Holmes} & {Video-MMMU (mc)} & {LongVideo-Reason} \\ 

\midrule
\rowcolor{StandardMLLMColor}
\multicolumn{9}{c}{\textit{Standard MLLMs}} \\
\rowcolor{StandardMLLMColor}
Qwen2.5-VL-7B  & 16 & {79.6} & 34.5 & 59.3 & \underline{64.8} & 36.4 & 55.2 & 69.8 \\
\rowcolor{StandardMLLMColor}
Qwen2.5-VL-7B (CoT) & 16 & - & - & - & 59.2$^\spadesuit$ & - & - & - \\
\rowcolor{StandardMLLMColor}
Video-R1-7B & 16 & 79.7 & \underline{35.7} & {61.4} & 64.2$^\spadesuit$ & \underline{40.4} & \underline{55.7} & {69.9} \\

\rowcolor{LatentMLLMColor}
\multicolumn{9}{c}{\textit{Latent MLLMs}} \\
\rowcolor{LatentMLLMColor}
Mull-Token-7B & 16 & \underline{80.0} & 35.3 & \textbf{62.7} & 64.0 & 38.4 & 51.3 & \underline{71.6} \\
\rowcolor{OurColor}
\textbf{VideoLatent-7B (Ours)} & 16 & \textbf{80.7} & \textbf{37.6} & \underline{61.7} & \textbf{65.8} & \textbf{40.5} & \textbf{57.7} & \textbf{71.7} \\
\rowcolor{OurColor}
\multicolumn{2}{c}{\textit{$\Delta$ (vs. Qwen2.5-VL-7B)}} & \textit{+1.1} & \textit{+3.1} & \textit{+2.4} & \textit{+1.0} & \textit{+4.1} & \textit{+2.5} & \textit{+1.9} \\

\bottomrule
\end{tabular}
}
\caption{\textbf{Additional experiments} on seven more benchmarks, covering general video understanding and complex video reasoning. Our model achieves the best or second-best performance on all benchmarks. 
$\spadesuit$: Results are borrowed from~\cite{videor1}, while others are produced under the same experimental setting.
(\ie, using 16 input frames and a resolution of 256$\times$32$\times$32 pixels).
\textbf{Bold} / \underline{Underlined}: Best / Second-best result.
}
\label{tab:additional_exp}
\end{table*}
\section{Experiments}

\subsection{Experimental Settings}
\noindent \textbf{Implementation Details.}
\label{sec:exp_setup}
In this work, our \model{} models are built upon Qwen2.5-VL~\cite{qwen25vl} and Qwen3-VL~\cite{qwen3vl}.
During model training, we strictly follow~\cite{videor1,videorft} to limit the maximum number of input video frames to 16 to ensure training efficiency, where each frame is processed at a max resolution of 128$\times$28$\times$28 pixels and 128$\times$32$\times$32 pixels for Qwen2.5-VL-based and Qwen3-VL-based models, respectively.
During evaluation, we further follow~\cite{videor1,videorft} to double the frame resolution (\eg, 256$\times$28$\times$28 pixels and 256$\times$32$\times$32 pixels for \model{}-7B and \model{}-8B, respectively) to facilitate a fair comparison. 
More implementation details are provided in the Appendix.

\noindent \textbf{Evaluation Benchmarks.}
To ensure a comprehensive evaluation, our experiments are conducted on 14 benchmarks, covering general video understanding and complex video reasoning. 
General video understanding comprises 3 short-video understanding benchmarks (\ie, MVBench~\cite{mvbench}, TempCompass~\cite{tempcompass}, Next-QA~\cite{nextqa}) and 4 long-video understanding benchmarks (\ie, Video-MME~\cite{videomme}, LongVideoBench~\cite{longvideobench}, LVBench~\cite{lvbench}, MLVU~\cite{lvbench}),
while complex video reasoning includes Video-TT~\cite{videott}, VCR-Bench~\cite{vcrbench}, VideoMathQA~\cite{videomathqa}, MMVU~\cite{mmvu}, Video-Holmes~\cite{Video-Holmes}, Video-MMMU~\cite{videommmu}, 
LongVideo-Reason~\cite{longvideoreason}.

\noindent \textbf{Comparison Baselines.}
To facilitate a thorough comparison, we compare our models against both standard and latent MLLMs.
Standard MLLMs comprise Qwen2.5-VL-3B/7B~\cite{qwen25vl}, Video-R1-7B~\cite{videor1}, TinyLLaVA-Video-R1~\cite{TinyLLaVA-Video-R1}, VideoRFT-7B~\cite{videorft}, 
Open-o3-Video-7B~\cite{open-o3-video}, 
VideoChat-R1.5~\cite{videochatr1.5}, Qwen3-VL-8B~\cite{qwen3vl}, OneThinker-8B~\cite{onethinker}, and VideoAuto-R1~\cite{videoautor1}.
Latent MLLMs encompass Monet-7B~\cite{monet}, LVR-7B~\cite{lvr}, and Mull-Token-7B~\cite{Mull-Tokens}.
We also include three closed-source models, including GPT-4o~\cite{gpt4o}, Gemini-2.5-Pro~\cite{gemini2.5}, and Seed1.5-VL~\cite{seed1.5}.

\subsection{Main Experimental Results}
To ensure a fair comparison, all models in Tab.~\ref{tab:main_exp} and Tab.~\ref{tab:additional_exp} are built upon Qwen2.5-VL-7B.

\smallskip
\noindent \textbf{Compared with standard MLLMs.} As shown in Tab.~\ref{tab:main_exp} and Tab.~\ref{tab:additional_exp}, our \model{}-7B consistently outperforms existing standard MLLMs across all 14 benchmarks under varying numbers of input frames (from 16 to 64).
For example, compared with the MLLM backbone, \model{}-7B exhibits stronger or comparable performance across all benchmarks (\eg, +3.7 on MVBench (short-video understanding; 16 frames), +2.7 on Video-MME (long-video understanding; 16 frames), and +6.5 on Video-TT (complex video reasoning; 16 frames)).
In contrast, we observe that the CoT-based MLLMs fail to achieve consistent performance improvement over the MLLM backbone.
For example, Video-R1-7B / Open-o3-Video-7B achieves -2.3 / -1.3 on MVBench (64 frames), -1.3 / -1.3 on Video-MME (64 frames), and -4.6 / -4.8 on VideoMathQA (64 frames).

\begin{table*}[!t]
\centering
\resizebox{\linewidth}{!}{
\setlength{\tabcolsep}{1.4mm}
\renewcommand\arraystretch{1.3}
\begin{tabular}{lcccccccc}
\toprule
\multirow{2}{*}{\textbf{Model}} & \multirow{2}{*}{\textbf{Frames}} & \multicolumn{4}{c}{\textbf{General Video Understanding}} & \multicolumn{3}{c}{\textbf{Complex Video Reasoning}} \\ 
\cmidrule(lr){3-6} \cmidrule(lr){7-9}
      & & {MVBench} & {TempCompass} & {Video-MME} & {LongVideoBench} & {Video-TT} & {VCR-Bench} & {VideoMathQA} \\ 
      
\midrule
\rowcolor{grey}
GPT-4o$^\spadesuit$ & - & - & - & 71.9 & 66.7 & 45.2 & 46.9 & 20.2 \\

\rowcolor{grey}
Gemini-2.5-Pro$^\spadesuit$ & - & - & - & 84.3 & - & - & - & - \\

\rowcolor{grey}
Seed1.5-VL$^\spadesuit$ & - & 74.4 & 83.7 & 77.9 & 74.0 & - & - & - \\

\rowcolor{grey}
VideoChat-R1.5$^\spadesuit$ & 2048 & 70.6 & - & 65.2 & 61.4 & - & - & - \\

\rowcolor{grey}
VideoAuto-R1$^\spadesuit$ & 2048 & 72.0 & - & 71.7 & 67.4 & - & - & - \\

\midrule
Qwen3-VL-8B & 16 & 67.7 & 75.6 & \textbf{60.7} & \textbf{58.3} & \underline{39.2} & \underline{49.4} & 29.0 \\

OneThinker-8B & 16 & \underline{68.4} & \textbf{76.6} & 57.4 & 56.3 & 38.7 & 46.9 & \underline{29.8} \\

\rowcolor{OurColor}
\textbf{VideoLatent-8B (Ours)} & 16 & \textbf{70.6} & \underline{76.5} & \underline{60.5} & \underline{57.8} & \textbf{44.4} & \textbf{51.0} & \textbf{31.4} \\

\rowcolor{OurColor}
\multicolumn{2}{c}{\textit{$\Delta$ (vs. Qwen3-VL-8B)}} & \textit{+2.9} & \textit{+0.9} & \textit{-0.2} & \textit{-0.5} & \textit{+5.2} & \textit{+1.6} & \textit{+2.4} \\

\midrule
Qwen3-VL-8B & 32 &  68.9 & 75.5 & \textbf{64.1} & \underline{60.1} & \underline{42.1} & \underline{53.9} & 30.0 \\

OneThinker-8B & 32 & \underline{69.9} & \textbf{76.3} & 61.4 & 59.6 & 39.2 & 53.1 & \underline{30.5} \\

\rowcolor{OurColor}
\textbf{VideoLatent-8B (Ours)} & 32 & \textbf{70.6} & \underline{76.2} & \underline{63.5} & \textbf{60.4} & \textbf{45.1} & \textbf{58.4} & \textbf{31.7} \\

\rowcolor{OurColor}
\multicolumn{2}{c}{\textit{$\Delta$ (vs. Qwen3-VL-8B)}} & \textit{+1.7} & \textit{+0.7} & \textit{-0.6} & \textit{+0.3} & \textit{+3.0} & \textit{+4.5} & \textit{+1.7} \\

\midrule
Qwen3-VL-8B & 64 & 69.0 & 75.7 & \textbf{67.2} & \underline{62.5} & \underline{43.2} & 55.3 & 30.2 \\

OneThinker-8B & 64 & \underline{69.9} & \underline{76.3} & 63.8 & 59.7 & 40.8 & \underline{55.7} & \textbf{32.6} \\

\rowcolor{OurColor}
\textbf{VideoLatent-8B (Ours)} & 64 & \textbf{71.6} & \textbf{76.4} & \underline{65.7} & \textbf{63.0} & \textbf{46.4} & \textbf{59.8} & \underline{31.4} \\

\rowcolor{OurColor}
\multicolumn{2}{c}{\textit{$\Delta$ (vs. Qwen3-VL-8B)}} & \textit{+2.6} & \textit{+0.7} & \textit{-1.5} & \textit{+0.5} & \textit{+3.2} & \textit{+4.5} & \textit{+1.2} \\

\bottomrule
\end{tabular}
}
\caption{\textbf{Generalization to a different MLLM backbone}. Our model improves the performance of Qwen3-VL-8B on most benchmarks, demonstrating its generalizability across architectures. 
$\spadesuit$: Results are borrowed from~\cite{onethinker,videott,vcrbench,videomathqa,seed1.5,videochatr1.5,videoautor1}, while others are produced under the same setting. 
\textbf{\textcolor{deepgrey}{Grey}}: Closed-source models or MLLMs using extremely large numbers of input frames.
\textbf{Bold} / \underline{Underlined}: Best / Second-best result. 
}
\label{tab:main_exp2}
\end{table*}

\begin{table*}[!t]
\centering
\resizebox{\linewidth}{!}{
\setlength{\tabcolsep}{1.4mm}
\renewcommand\arraystretch{1.3}
\begin{tabular}{lcccccccc}
\toprule
\multirow{2}{*}{\textbf{Model}} & \multirow{2}{*}{\textbf{Frames}} & \multicolumn{4}{c}{\textbf{General Video Understanding}} & \multicolumn{3}{c}{\textbf{Complex Video Reasoning}} \\ 
\cmidrule(lr){3-6} \cmidrule(lr){7-9}
      & & {MVBench} & {TempCompass} & {Video-MME} & {LongVideoBench} & {Video-TT} & {VCR-Bench} & {VideoMathQA} \\ 

\midrule
Qwen2.5-VL-3B  & 16 & \underline{62.9} & \underline{68.0} & \underline{53.9} & \underline{52.4} & \underline{38.7} & \underline{44.7} & \underline{24.8} \\
TinyLLaVA-Video-R1$^\spadesuit$ & 16 & 49.5 &  - & 46.6 & - & - & - & - \\
VideoRFT-3B$^\spadesuit$ & 32 & 59.5 & 61.0 & 45.4 & - & - & - & - \\
\rowcolor{OurColor}
\textbf{VideoLatent-3B (Ours)} & 16 & \textbf{65.2} & \textbf{69.8} & \textbf{54.4} & \textbf{54.2} & \textbf{44.6} & \textbf{45.5} & \textbf{25.7} \\

\rowcolor{OurColor} 
\multicolumn{2}{c}{\textit{$\Delta$ (vs. Qwen2.5-VL-3B)}} & +2.3 & +1.8 & +0.5 & +1.8 & +5.9 & +0.8 & +0.9 \\

\bottomrule
\end{tabular}
}
\caption{\textbf{Generalization to a small-scale model}. Our model outperforms other models, further demonstrating its effectiveness across different model scales.
$\spadesuit$: Results are borrowed from~\cite{TinyLLaVA-Video-R1,videorft}, while others are produced under the same experimental setting. 
\textbf{Bold} / \underline{Underlined}: Best / Second-best result. 
}
\label{tab:exp_3b}
\end{table*}

Moreover, our \model{}-7B surpasses the CoT-based MLLMs by a notable margin on most benchmarks.
For example, compared with Video-R1-7B / Open-o3-Video-7B, \model{}-7B achieves 
+5.5 / +3.4 on MVBench (32 frames), 
+2.4 / +2.4 on Video-MME (64 frames),
and +5.7 / +4.3 on VideoMathQA (32 frames).

\smallskip
\noindent \textbf{Compared with latent MLLMs.}
Tab.~\ref{tab:main_exp} and Tab.~\ref{tab:additional_exp} further demonstrate that our \model{}-7B outperforms latent MLLMs on most benchmarks.
For Monet-7B, we observe that it often struggles to follow instructions, leading to relatively weak performance on video-language tasks.
Compared with LVR-7B / Mull-Token-7B, our \model{}-7B achieves +3.7 / +1.3 on MVBench (short-video understanding; 16 frames), +0.9 / +1.6 on LongVideoBench (long-video understanding; 16 frames), and +3.6 / +4.1 on VideoMathQA (complex video reasoning; 16 frames).

\smallskip
\noindent\textbf{Overall.} These extensive comparative experiments validate the effectiveness of our model design and the proposed latent self-forcing for enhancing video understanding and reasoning.

\begin{table*}[t]
\centering
\resizebox{\linewidth}{!}{
\setlength{\tabcolsep}{1.4mm}
\renewcommand\arraystretch{1.3}
\begin{tabular}{lcccccccc} 
\toprule
\multirow{2}{*}{\textbf{Model}} & \multirow{2}{*}{\textbf{Frames}} & \multicolumn{4}{c}{\textbf{General Video Understanding}} & \multicolumn{3}{c}{\textbf{Complex Video Reasoning}} \\ 
\cmidrule(lr){3-6} \cmidrule(lr){7-9}
      & & {MVBench} & {TempCompass} & {Video-MME} & {LongVideoBench} & {Video-TT} & {VCR-Bench} & {VideoMathQA} \\ 

\midrule
SFT & 16 & {65.8} & {72.4} & {57.5} & {53.0} & \textbf{43.9} & {46.3} & {28.8} \\ 

\rowcolor{OurColor}
\textbf{Latent Self-Forcing (Ours)} & 16 & \textbf{68.3} & \textbf{73.7} & \textbf{59.3} & \textbf{57.3} & {43.4} & \textbf{49.6} & \textbf{30.5} \\

\rowcolor{OurColor}
\multicolumn{2}{c}{\textit{$\Delta$ (vs. SFT)}} & \textit{+2.5} & \textit{+1.3} & \textit{+1.8} & \textit{+4.3} & \textit{-0.5} & \textit{+3.3} & \textit{+1.7} \\

\bottomrule
\end{tabular}
}
\caption{
\textbf{Comparison between different training methods}. Our latent self-forcing consistently outperforms standard SFT on most benchmarks.
\textbf{Bold}: Best result.
}
\label{tab:ablation_sft}
\end{table*}
\begin{figure*}[t]
    \centering
    \includegraphics[width=\linewidth]{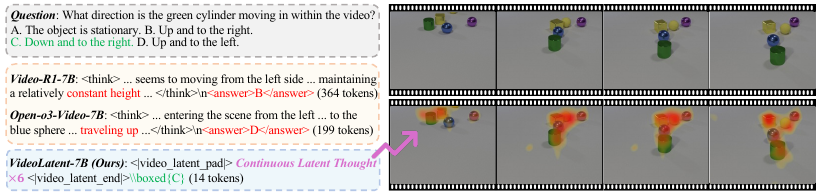}
    \caption{Case study and heatmap visualization of the generated latent thoughts. See more details in Appendix.}
    \label{fig:visualization}
\end{figure*}

\begin{table*}[t]
\centering
\resizebox{0.65\linewidth}{!}{
\setlength{\tabcolsep}{1.4mm}
\begin{tabular}{cccc >{\columncolor{OurColor}}c c}
\toprule
\textbf{$N_{\text{max}}$} & 0 & 2 & 4 & 6 & 8 \\ 
\midrule
\textbf{LongVideoBench} (16 frames) & 55.4 & 56.3 & 56.7 & \textbf{57.3} & 56.4 \\
\bottomrule
\end{tabular}
}
\caption{
Ablation of latent reasoning steps $N_{\text{max}}$.
\textbf{Bold}: Best result.
}
\label{tab:ablation_latent_szie}
\end{table*}

\begin{table*}[ht]
    \centering
    \resizebox{\linewidth}{!}{
    \setlength{\tabcolsep}{1.4mm}
    \begin{tabular}{l cccc ccc >{\columncolor{OurColor}}c }
    \toprule
    & \multicolumn{4}{c}{\textbf{Latent Self-Forcing}} & \multicolumn{3}{c}{\textbf{Latent Injection Module}} &  \\ 
    \cmidrule(lr){2-5} \cmidrule(lr){6-8}
    \multirow{-2}{*}{\textbf{Model}} & w/o $\mathcal{L}_{\text{lv}}$ & w/o $\mathcal{L}_{\text{lq}}$ & w/o $\mathcal{L}_{\text{inter}}$ & w/o $\mathcal{L}_{\text{intra}}$ & w/o $\mathbf{c}_{\text{z}v}$ & w/o $\mathbf{c}_{\text{z}q}$ & w/o $\mathbf{c}_{qv}$ & \multirow{-2}{*}{\textbf{Full}} \\ 
    \midrule
    \textbf{LongVideoBench (16 frames)}  & 55.6 & 56.1 & 56.6 & 56.6 & 55.6 & 55.2 & 55.3 & \textbf{57.3} \\ 
    \bottomrule
    \end{tabular}
    }
\caption{
Ablation of training and module designs. 
\textbf{Bold}: Best result.
}
\label{tab:ablation_loss}
\end{table*}

\subsection{Generalization Analysis}

\noindent \textbf{Generalization to different MLLM backbones.}
To validate the generalizability of our approach to different MLLM backbones, we extend our method to Qwen3-VL-8B (using $N_{\text{max}}=2$ to reduce training overhead). 
Tab.~\ref{tab:main_exp2} shows that \model{}-8B consistently yields superior or competitive performance against the standard MLLMs under the same experimental setting.
Specifically, compared with the MLLM backbone / OneThinker-8B, our \model{}-8B achieves +2.9 / +2.2 on MVBench (short-video understanding; 16 frames), +0.5 / +3.3 on Video-MME (long-video understanding; 64 frames), +5.2 / +5.7 on Video-TT (complex video reasoning; 16 frames).
Moreover, 
our \model{}-8B (64 frames) even surpasses VideoChat-R1.5 (using extremely large numbers of input frames) on general video understanding (\eg, +1.0 on MVBench, +0.5 on Video-MME, and +1.6 on LongVideoBench), 
while also outperforming GPT-4o on some complex video reasoning tasks (\eg, +1.2 on Video-TT and +12.9 on VCR-Bench).

\smallskip
\noindent \textbf{Generalization to different model scales.}
To assess the generalizability of our approach under limited computational resources, we further extend our method to Qwen-2.5VL-3B (using $N_{\text{max}}=2$ to reduce training overhead).
Tab.~\ref{tab:exp_3b} shows that our \model{}-3B achieves the best performance across all benchmarks, outperforming both the MLLM backbone and prior works with more input frames (\ie, VideoRFT-3B~\cite{videorft}).

\smallskip
\noindent \textbf{Overall.}
These results demonstrate the strong generalizability of our method across different MLLM backbones and model scales.

\subsection{Further Analyses}
\noindent \textbf{SFT vs. Latent self-forcing.} Tab.~\ref{tab:ablation_sft} shows that our latent self-forcing consistently outperforms standard SFT on most benchmarks, validating the effectiveness of the proposed method.

\smallskip
\noindent \textbf{Computational efficiency.}
Fig.~\ref{fig:overhead} demonstrates that our method exhibits significantly improved computational efficiency (training overheads are borrowed from~\cite{videor1,onethinker}, inference overheads are calculated based on VideoMathQA (64 frames).
Specifically, compared with Video-R1-7B, \model{}-7B significantly reduces training overhead by $\sim$6$\times$ and inference overhead by $\sim$68$\times$, while maintaining strong performance (as shown in Tab.~\ref{tab:main_exp} and Tab.~\ref{tab:additional_exp}).

\smallskip
\noindent \textbf{Qualitative analysis.}
Fig.~\ref{fig:visualization} presents qualitative comparisons between CoT-based MLLMs and our model.
The case study shows that, while CoT-based MLLMs exhibit overthinking behaviors and incorrect reasoning, our model effectively internalizes the video reasoning with significant inference efficiency, consistent with the results in Fig.~\ref{fig:overhead}.

Moreover, to better understand the implicit reasoning process, we provide a heatmap visualization of the self-generated continuous latent thoughts based on the similarity between video patch tokens and latent thoughts.
Specifically, we observe that the latent thoughts dynamically capture key objects across different frames, thereby enhancing video understanding and reasoning capabilities.

\subsection{Ablation Studies}
\noindent \textbf{Latent reasoning steps.}
As shown in Tab.~\ref{tab:ablation_latent_szie}, we ablate the effect of the maximum number of latent reasoning steps (\ie, $N_{\text{max}}$ in Sec.~\ref{sec:our_model}). 
The results demonstrate that $N_{\text{max}}=6$ achieves the best performance, while even $N_{\text{max}}=2$ outperforms $N_{\text{max}}=0$ (\ie, the MLLM backbone). 
The performance first increases and then decreases as $N_{\text{max}}$ increases, suggesting that an overly small $N_{\text{max}}$ is insufficient for effective video-language learning, while an excessively large $N_{\text{max}}$ may impair the backbone's pretrained text reasoning capability.

\smallskip
\noindent \textbf{Latent self-forcing.}
Tab~\ref{tab:ablation_loss} presents ablations of different latent-centric objectives, where removing any loss consistently degrades performance, demonstrating the effectiveness of our latent self-forcing training paradigm.
Among these objectives, $\mathcal{L}_{\text{lv}}$ and $\mathcal{L}_{\text{lq}}$ contribute the most (+1.7 and +1.2), highlighting the importance of aligning latent thoughts with both video and question contexts.

\smallskip
\noindent \textbf{Latent injection module.}
Tab~\ref{tab:ablation_loss} further ablates the design of the latent injection module, 
where removing any injected component consistently leads to performance degradation,
indicating the importance of injecting both video and question information for enhancing video-language learning.

\section{Conclusion}
In this paper, we present \model{}, a novel framework specially designed for visual latent reasoning on video understanding and reasoning.
Specifically, \model{} involves a new latent injection module and latent self-forcing training paradigm, enabling effective latent reasoning without relying on additional supervision signals.
Extensive experiments conducted on fourteen video-language benchmarks demonstrate that our \model{} consistently outperforms existing standard and latent MLLMs across general video understanding and complex video reasoning, while achieving superior efficiency and generalizability.

\section*{Limitations}
We summarize the limitations as follows:
(1) Despite our model achieving strong performance across various general video understanding and complex video reasoning benchmarks, the results may not generalize to all video-language benchmarks; 
(2) Due to computational constraints, the maximum number of input video frames in our current training is limited to 16.
Future work could benefit from leveraging higher-quality training data with an increased number of frames to further enhance video-language learning;
(3) Since our method relies on self-generated latent thoughts without additional supervision signals, the generated latent thoughts may still be irrelevant to the video and question context;
and (4) Despite its superior inference efficiency, similar to prior visual latent reasoning works, our implicit latent reasoning may provide relatively lower interpretability compared with explicit CoT reasoning.

\bibliography{custom}

\newpage
\appendix
In the Appendix, we provide more implementation details in Sec.~\ref{appendix:implementation}, including MLLM backbones, training details, training time cost, evaluation details, and instruction prompt. 
Subsequently, we describe the statistics of both training data and evaluation data in Sec.~\ref{appendix:statistics}.
Finally, we provide comprehensive visualization results in Sec.~\ref{appendix:visualization}.

\section{Implementation details}
\label{appendix:implementation}
\paragraph{MLLM Backbones.}
In this work, our \model{}-7B, \model{}-8B models, and \model{}-3B are built upon Qwen2.5-VL-7B-Instruct~\cite{qwen25vl}, Qwen3-VL-8B-Instruct~\cite{qwen3vl}, and Qwen2.5-VL-3B-Instruct~\cite{qwen25vl}, respectively.
Our main experiments are conducted on \model{}-7B (\eg, Tab.~\ref{tab:main_exp}, Tab.~\ref{tab:additional_exp}, Tab.~\ref{tab:ablation_sft}, Tab.~\ref{tab:ablation_loss}, Fig.~\ref{fig:radar}, Fig.~\ref{fig:overhead}, and Fig.~\ref{fig:visualization}), while we validate the generalizability of our method using \model{}-8B models (\ie, Tab.~\ref{tab:main_exp2}) and \model{}-3B (\ie, Tab.~\ref{tab:exp_3b}).

\paragraph{Training details.}
As mentioned in the main paper, for training, we strictly follow~\cite{videor1,videorft} to limit the maximum number of input video frames to 16 to ensure training efficiency.
Specifically, the maximum resolution of each frame is set to 128$\times$28$\times$28 pixels and 128$\times$32$\times$32 pixels for Qwen2.5-VL-based and Qwen3-VL-based models, respectively. 

We train our models for 2 epochs with a per-device batch size of 1 and gradient accumulation steps of 16 on 4 NVIDIA H800 (80GB) GPUs, resulting in a total batch size of 64. 
During training, the modality projector, the LLM, and the proposed latent injection module are trainable, while the vision encoder remains frozen.
To ensure training stability, we use the Adam optimizer with a learning rate of 1e-6, while employing a learning rate of 1e-5 on the newly introduced latent injection module.
We further apply a linear decay scheduler with a warmup of 100 steps, a weight decay of 0.01, and a maximum gradient norm of 1.0.

For the latent self-forcing training paradigm, we set the maximum number of latent reasoning steps $N_{max} = 6$ for our \model{}-7B to achieve the best performance (as demonstrated in Tab.~\ref{tab:ablation_latent_szie}), while using $N_{max} = 2$ for \model{}-3B and \model{}-8B to reduce training overhead.
The queue size of the memory banks and temperature hyperparameter used for contrastive learning (Eq.~\ref{eq:loss_lv}, Eq.~\ref{eq:loss_lq}, and Eq.~\ref{eq:loss_inter}) are set to $K=64$ and $\tau=0.2$, respectively.
The balancing coefficients in Eq.~\ref{eq:loss_total} are set to $\lambda_{\text{cl}}=0.1$ and $\lambda_{\text{intra}}=0.01$.

\paragraph{Training time cost.}
For training, we use  with DeepSpeed Zero2~\cite{deepspeed} to accelerate the training, where \model{}-7B ($N_{max} = 6$) takes 17 hours (\ie, 68 H800 GPU hours), \model{}-8B ($N_{max} = 2$) takes 14 hours (\ie, 64 H800 GPU hours), and \model{}-3B ($N_{max} = 2$) takes 8 hours (\ie, 32 H800 GPU hours).

\paragraph{Evaluation details.}
During evaluation, we further follow~\cite{videor1,videorft} to double the frame resolution (\eg, 256$\times$28$\times$28 pixels and 256$\times$32$\times$32 pixels for \model{}-7B and \model{}-8B, respectively) to facilitate a fair comparison. 
All experiments are conducted with 1 NVIDIA H800 GPU.

\paragraph{Instruction prompt.}
We use the following instruction prompt to evaluate our \model{}:
\texttt{Please think with the given video and answer the multiple-choice question with the option's letter.\textbackslash nThe final answer MUST BE put in \textbackslash\textbackslash boxed\{\}.}
When reproducing the experimental results of the baselines, we use their official system and instruction prompts provided in their paper or official code.
For prior visual latent reasoning works, we adopt the multi-image setting to evaluate the performance of Monet-7B~\cite{monet}and LVR-7B~\cite{lvr}, while Mull-Token-7B directly takes video inputs.


\section{Data Statistics}
\label{appendix:statistics}
In this section, we provide the statistics of both our training data and evaluation data.

\paragraph{Training data.}
To ensure reproducibility, our training is conducted on publicly available data, containing 81k video-question-answer triplets without any additional annotations (\eg, CoT traces, bounding boxes). The detailed statistics of the training data are listed in Tab.~\ref{tab:train_data_statistics}.

\begin{table}[t]
\centering
\resizebox{\linewidth}{!}{
\setlength{\tabcolsep}{1.4mm}
\begin{tabular}{l c}
\toprule
\textbf{Dataset} & \textbf{\#Samples} \\
\midrule
LLaVA-Video-178K~\cite{LLaVA-Video} & 59,641 \\
NeXT-QA~\cite{nextqa} & 6,095 \\
PerceptionTest~\cite{PerceptionTest} & 3,857 \\
CLEVRER~\cite{CLEVRER} & 4,188 \\
STAR~\cite{STAR} & 6,942 \\
\midrule
\textbf{Total} & \textbf{80,723} \\
\bottomrule
\end{tabular}
}
\caption{Statistics of the training data.}
\label{tab:train_data_statistics}
\end{table}

\begin{table}[t]
\centering
\resizebox{\linewidth}{!}{
\setlength{\tabcolsep}{1.4mm}
\begin{tabular}{l c}
\toprule
\textbf{Dataset} & \textbf{\#Samples} \\
\midrule
\multicolumn{2}{l}{\textit{General Video Understanding (Short Video)}} \\
MVBench~\cite{mvbench} & 4,000 \\
TempCompass~\cite{tempcompass} & 7,540 \\
Next-QA~\cite{nextqa} & 8564 \\
\midrule
\multicolumn{2}{l}{\textit{General Video Understanding (Long Video}} \\
Video-MME~\cite{videomme} & 2,700 \\
LongVideoBench~\cite{longvideobench} & 1,337 \\
LVBench~\cite{lvbench} & 1549 \\
MLVU~\cite{lvbench} & 2174 \\
\midrule
\multicolumn{2}{l}{\textit{Complex Video Reasoning}} \\
Video-TT~\cite{videott} & 1,000 \\
VCR-Bench (mc)~\cite{vcrbench} & 510 \\
VideoMathQA (mc)~\cite{videomathqa} & 420 \\
MMVU (mc)~\cite{mmvu} & 625 \\
Video-Holmes~\cite{Video-Holmes} & 1837 \\
Video-MMMU (mc)~\cite{videommmu} & 600 \\
LongVideo-Reason~\cite{longvideoreason} & 1,000 \\
\midrule
\textbf{Total} & \textbf{33,856} \\
\bottomrule
\end{tabular}
}
\caption{Statistics of the evaluation data, where some data splits are borrowed from Video-R1~\cite{videor1} and OneThinker~\cite{onethinker}.}
\label{tab:eval_data_statistics}
\end{table}

\paragraph{Evaluation data.}
In the main paper, we conduct extensive experiments on 14 video-language benchmarks, covering 7 general video understanding benchmarks, and 7 complex video reasoning benchmarks.
The detailed statistics of the evaluation data are listed in Tab.~\ref{tab:eval_data_statistics}. 

Our main experiments (\eg, Tab.~\ref{tab:main_exp}, Tab.~\ref{tab:main_exp2}, and Tab.~\ref{tab:exp_3b}) are conducted on seven widely used video-language benchmarks due to computational constraints.
For general video understanding, we evaluate on five benchmarks, comprising 
two short-video benchmarks (\ie, MVBench~\cite{mvbench} and TempCompass~\cite{tempcompass}) 
and two long-video benchmarks (\ie, Video-MME~\cite{videomme} and LongVideoBench~\cite{longvideobench}).
For complex video reasoning, we evaluate on three benchmarks, including VCR-Bench~\cite{vcrbench}, VideoMathQA~\cite{videomathqa}, and Video-TT~\cite{videott}.

Additionally,  we provide experimental results on seven more benchmarks (\ie, Next-QA~\cite{nextqa}, LVBench~\cite{lvbench}, MLVU~\cite{mlvu}, MMVU~\cite{mmvu}, Video-MMMU~\cite{videommmu}, Video-Holmes~\cite{Video-Holmes}, and LongVideo-Reason~\cite{longvideoreason}) in Tab.~\ref{tab:additional_exp} to further validate the effectiveness of our method.

\section{Visualizations}
\label{appendix:visualization}
\begin{figure*}[t]
    \centering
    \includegraphics[width=0.95\linewidth]{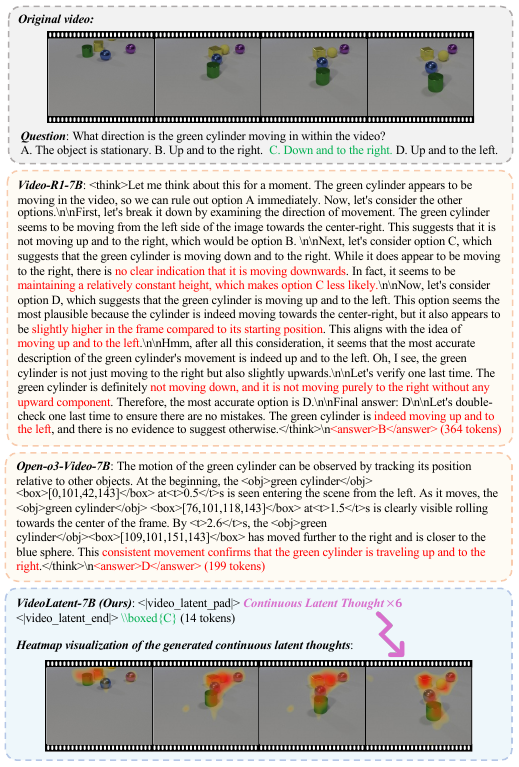}
    \caption{Comprehensive visualization result of a case study and heatmap visualization of the generated continuous latent thought, which is the detailed version of Fig.~\ref{fig:visualization} presented in the main paper.}
    \label{fig:appendix_visualization1}
\end{figure*}

\begin{figure*}[t]
    \centering
    \includegraphics[width=0.95\linewidth]{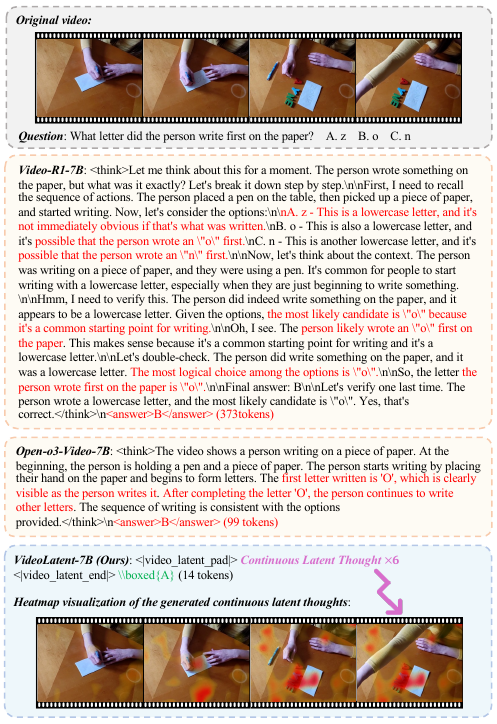}
    \caption{Comprehensive visualization result of a case study and heatmap visualization of the generated continuous latent thoughts.}
    \label{fig:appendix_visualization2}
\end{figure*}

In this section, we provide comprehensive case studies and heatmap visualizations of the self-generated latent thoughts in Fig.~\ref{fig:appendix_visualization1} and Fig.~\ref{fig:appendix_visualization2}.


\end{document}